\documentclass[10pt, twocolumn]{article}
\usepackage[utf8]{inputenc}
\usepackage{geometry}
\usepackage{graphicx}
\usepackage{amsmath,amsfonts,amssymb}
\usepackage[sort,numbers]{natbib}
\usepackage{fancyhdr}
\usepackage{titlesec}
\usepackage{caption}
\usepackage{hyperref}

\titleformat{\title}{\normalfont\Large\bfseries}{\thesection}{1em}{}
\titlespacing*{\title}{0pt}{\dimexpr -\baselineskip+1in}{2em}

\title{
    \vspace{-2.0em}  
    \textbf{A Survey of Learning with Imbalanced Data, Representation Learning and
SEP Forecasting} \\
    \vspace{1.0em}  
    \large Josias Moukpe \\
    \vspace{0.5em}  
    \small Florida Institute of Technology \\
    \vspace{0.5em}  
    \small Department of Computer Science and Engineering \\
    \vspace{0.5em}  
}
\author{}
\date{}

\begin{document}

\maketitle
\thispagestyle{fancy}  

\begin{abstract}

Deep Learning methods have significantly advanced various data-driven tasks such as regression, classification, and forecasting. However, much of this progress has been predicated on the strong but often unrealistic assumption that training datasets are balanced with respect to the targets they contain. This misalignment with real-world conditions, where data is frequently imbalanced, hampers the effectiveness of such models in practical applications. Methods that reconsider that assumption and tackle real-world imbalances have begun to emerge and explore avenues to address this challenge. One such promising avenue is representation learning, which enables models to capture complex data characteristics and generalize better to minority classes. By focusing on a richer representation of the feature space, these techniques hold the potential to mitigate the impact of data imbalance. In this survey, we present deep learning works that step away from the balanced-data assumption, employing strategies like representation learning to better approximate real-world imbalances. We also highlight a critical application in SEP forecasting where addressing data imbalance is paramount for success.
\end{abstract}

\section{Introduction}
Deep Learning techniques have achieved significant advancements across a variety of tasks, including but not limited to regression, and classification. These tasks are useful for many more applications such as image recognition, object detection, weather prediction, speech analysis, text generation, and recommendation systems just to name a few. All these applications are data-driven where a dataset or collection of experiences is gathered to train a model that learns to accomplish the task required for the application. The model does better for the task and application when the quantity and the quality of the data increase. However, there is a gap between the kind of data these models are trained on and the kind of experiences these models will deal with in real-world applications. Models are trained in labs under the assumption that all the targets, whether classes in classification or range of values in regression, are balanced i.e. the number of instances in the data for these targets is roughly the same. We say that they are assuming those targets to be uniformly distributed. The gap lies in the fact that this strong assumption does not always hold in real-world applications. For example, let's consider weather forecasting where we are interested in training a model to predict if it's going to storm based on collected data of hours of weather activity. In this example, because there are so few storms compared to sunny or rainy hours, most of the data will feature sunny or rainy hours and a few stormy hours. Deep Learning methods, being statistical in nature, with the assumption that sunny, rainy, and stormy hours are all equally distributed, will mostly and wrongly predict storms never occurring. This breaks the assumption of target balance in the data and we say that the collected data of weather activity is imbalanced where storms are rare while sunshine and rain are frequent. Deep Learning methods built on this assumption fail to perform well on the desired task. To tackle this, researchers have started to drop these assumptions about the data and develop methods that handle the imbalanced nature of real-world data. It is also worth noting that some works use the term long-tail distribution to refer to imbalanced distributions. In this context, representation learning emerges as a pivotal approach. By learning richer, more nuanced representations of data, Deep Neural Networks become inherently robust to imbalanced distributions. This is achieved by enabling the model to understand and generalize complex features across both majority and minority targets, thereby mitigating the inherent biases caused by imbalanced data.\\ 
In this survey, we will feature some of those recent works that deal with imbalances in classification, regression, and forecasting tasks. We will also present methods that focus on learning good semantic representations in Deep Neural Networks which make them inherently robust to imbalance. Finally, we will present Solar Energetic Particle (space weather) classification, regression, and forecasting as a challenging real-world application where these issues are further exacerbated, leaving an open door of opportunity for major improvements.

\begin{figure*}[t!]  %
    \includegraphics[width=\textwidth]{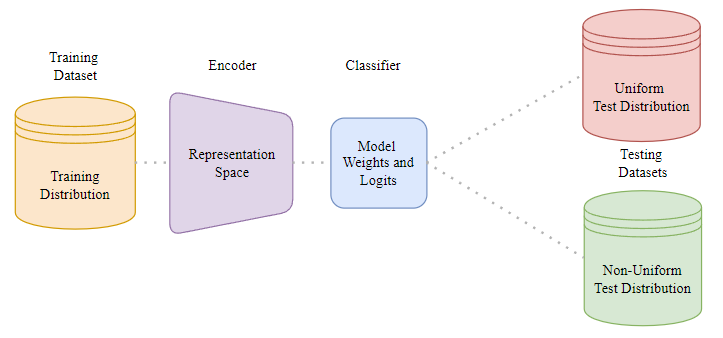} 
    \caption{Placement of the Imbalanced Classification groups in this section according to where they fit in the optimization process. The "Training Distribution" methods focus on the training dataset. The "Representation Space" methods focus on the encoder neural network and its encoded representations. The "Model Weights and Logits" methods focus on the parameters and output of the classifier model respectively. The "Non-Uniform Test Distribution" methods focus on the test dataset.}
    \label{fig:imbal-class}
\end{figure*}

\section{Imbalanced Classification}
For classification tasks, the imbalance is found with the classes of instances within the dataset. In imbalanced classification, the dataset is composed of several classes with some of those classes having orders of magnitude fewer samples than others. Those classes with fewer samples are called rare, infrequent, minority, or tail classes while the classes with more samples are called frequent, majority, or head classes. The imbalanced distribution is called long-tail distribution following a long-tail curve when the number of samples per class is plotted against the class index in sorted descending order. When naively training a classifier neural network on an imbalanced classification dataset, the network mostly considers the frequent or head class samples, as they dominate the training, and barely notices the rare or tail class samples. This leads to poor performance of the network on those tail classes and thus on the desired application. To tackle this issue, several works have been produced in recent years, addressing imbalance classification from different angles. Below we list the chosen angles and group the works under them. Figure~\ref{fig:imbal-class} illustrates where those angles fit in the imbalanced classification.

\subsection{Training Distribution}
To address imbalanced classification, some works focus on the input data distribution the model is trained on. This section is the far left side displayed in Figure~\ref{fig:imbal-class}. These works acknowledge the input distribution over the class samples is going to be imbalanced and develop techniques to re-balance it. 

Kang et al. \cite{kang2020decoupling}. proposed decoupling representation learning and classifier learning for long-tail data distributions, unlike previous methods which would train a model jointly for both representation and classifier learning. One advantage of decoupling that they point out is the ability to use different sampling techniques for the separate training stages and experimentally observe which sample technique benefits each stage the best. For their experiments, kang et al. set up a Convolutional Neural Network (CNN) which is composed of a backbone for representation learning and a classifier head, either as a Linear Model or a Multi-Layer Perception. They then employ various sampling techniques for both parts independently. For learning representations, they used \textbf{Instance balanced sampling} where each training example has an equal probability of being sampled resulting in classes with a greater number of samples dominating the training process. Once the representations are learned, to train the classifier, they proposed \textbf{cRT or Classifier Re-training}  where the classifier head is retrained on class-balanced sampling where a class is uniformly selected from the set of all classes and then the sample is uniformly selected from the selected class. After the test on benchmark long-tail datasets, Kang et al. show that data imbalance is not an issue for learning high-quality representations. Moreover, representation learning is not affected by the tail distribution but rather takes advantage of it. However, classifier learning is affected by the tail distribution and therefore needs to be balanced. The best coupling they found was instance-balanced sampling for representation learning and some balancing for classifier learning (cRT).

Still, within the framework of re-balancing the input distribution, Zhou et al.\cite{zhou2020bbn} proposed a different approach from the 2-stage decoupling of Kang et al. \cite{kang2020decoupling}. First, similarly to Kang et al., \cite{kang2020decoupling}, Zhou et al.\cite{zhou2020bbn} observed that balancing methods (when performed jointly) promote minority classes at the expense of the majority classes. They show through their experiments that for representation learning, using the original long-tail distribution has the lowest error rate overall. However, for classifier learning, using a balanced distribution has the least error rate overall. In an effort to reconcile the two, they propose a bilateral-branch network (BBN) that has two branches: one branch called the conventional branch that focuses on learning on the original imbalanced distribution for strong representation learning, and the other branch called the re-balancing branch that focuses on learning on a reverse distribution, where the rarer the class, the more likely it is to be sampled, for strong classifier learning. It's crucial to note that both branches share the same weights for the representation learning backbone. Both branches would then be combined through a cumulative learning strategy. The cumulative learning strategy features two classifiers from both branches. A classifier from the conventional branch to classify majority classes and a classifier from the re-balancing branch to classify minority classes. The cumulative learning strategy is a weighted sum of the two branches' outputs. That weight is modulated by the iteration number and follows a parabolic decay. Early in the training, more weight is given to the conventional branch while much later in the training, more weight is given to the re-balancing branch. The authors also propose a weighted cross-entropy loss that accompanies the cumulative learning strategy. During inference, both branches are considered equally and the final prediction is the average of the two branches' predictions. Their results show that their soft decoupling by transitioning from conventional learning to re-balancing learning is better than the prior techniques. The norms of the classifier weights are also shown to be more balanced, with a lower standard deviation, than the prior techniques.

Another approach to balancing the input distribution is to generate synthetic samples. Ren et al.\cite{wang2021rsg} propose a simple but effective module called RSG to address the long-tail problem. RSG works by generating new samples for the rare classes. RSG uses variation information among the real samples from the frequent classes to generate new samples for the rare classes. RSG is composed of 3 sub-modules:
\begin{itemize}
    \item \textbf{Center estimation module} finds class centers
    \item \textbf{Contrastive module} checks if two feature maps belong to the same class
    \item \textbf{Vector Transformation module} generates new rare samples using displacement vectors
\end{itemize}
The center estimation module and the contrastive module are both trained using the center estimation with sample contrastive loss and the vector transformation module is trained using a maximized vector loss. Injected between layers of a CNN, the RSG module takes the feature maps from the previous layer and outputs new synthetic samples for the rare classes to be added to the existing set of samples. The center estimation module (implemented as a linear model) estimates takes the feature maps and estimates a set of centers in each class which can be used as an anchor for obtaining the feature displacement vector of each sample. The contrastive module ensures that no information relevant to frequent classes is present in the feature displacement vector. It is implemented with a CNN that will output a probability distribution of yes/no that 2 feature maps belong to the same class. The vector transformation module calculates the feature displacement of each frequent-class sample based on estimated centers and uses the displacement to generate new samples for the rare classes. First, a displacement vector is calculated for each sample by subtracting the sample's feature map from its closest up-sampled center. Then the displacement vector is transformed by a convolutional layer to produce a displacement vector that will be applied to the rare-class sample as opposed to the center to push away the decision boundary and enlarge the feature space. The transformed vector is optimized by a maximized vector loss to be co-linear with the rare-class displacement vector to its center. The loss function is a linear combination of the center estimation with sample contrastive loss, the maximized vector loss, and the classification loss.

Another approach to balancing the input distribution was explored by Li et al. \cite{li2021ssd} where the authors aimed at balancing the input distribution by using soft labels. Unlike hard labels which indicate a 100\% probability (certainty) of a sample belonging to a class, soft labels encode uncertainty by assigning to a sample a probability distribution over the classes. Soft labels allow frequent classes to boost the effective sample count of rare classes. They propose a multi-stage training process. The first stage is to learn a good feature extractor with a classification head and self-supervision head. The classification head learns to classify the input images into the classes under imbalanced sampling. The self-supervision head learns to predict the rotation of the input images and contrast an instance input from other instances. Once trained, the second stage uses the frozen weights of the feature extractor and trains a classifier head with balanced sampling. The learned network at this second stage is considered the teacher network, which will be used the produce soft labels on the dataset. Then the third stage is to train a student backbone with a classifier head that is trained on the hard labels and a self-distillation head that is trained on the soft labels, both using imbalanced sampling. The classifier head only is used at the end for inference. Optionally the classifier head can be further trained using class-balanced sampling in a fourth stage.

All these works address the imbalance classification from the angle of re-balancing the distribution of the input data used for training the model. The next set of works takes a different angle and addresses the imbalance in the model itself during training.

\subsection{Model Logits and Weights}
To address imbalanced classification, these works observed that the optimization procedure of the model can be adjusted to handle better rare classes. This section is illustrated by the blue square in Figure~\ref{fig:imbal-class}.

Kang et al. \cite{kang2020decoupling}, from the previous section also belong in this category as they proposed 
\begin{itemize}
    \item \textbf{$\tau$-normalized classifier} where the norms of the classifier weights associated with each class are normalized to prevent the majority classes' weight from growing dominant as they normally would during training
    \item \textbf{LWS or Learnable Weights Scaling} takes that step further than $\tau$-normalized and makes the normalization coefficient of the class-associated weights learnable
\end{itemize}
Their $\tau$-normalization and LWS techniques adjust the weights of the model to achieve class re-balancing for the second stage of the training where a classifier is trained.

Unlike Kang et al. \cite{kang2020decoupling}, Ren et al.\cite{ren2020balsoftmax} observed that the softmax function is not well suited for long-tail data distributions as it gives bias gradient estimates under long-tail data distributions. They therefore propose a balanced softmax function for use only during training while the conventional softmax function is used during testing or inference (in deployment). Utilizing the Bayesian definition of the softmax function, Their balanced softmax function is softmax where the number of samples in each class is a coefficient to the exponential terms. This formulation of a balanced softmax during training forces the training network to output larger logits for minority classes. However, paired with a class-balanced strategy, Ren et al. observed that minority classes were overbalanced. They needed a new sampling strategy that would work with the balanced softmax function. They proposed a meta-sampler, an additional model that could learn the optimal sampling strategy for the balanced softmax function during training. The training routine would be in two levels, the first level would be the training of the meta sampler in the inner training loop and then the training of the desired classification model in the outer training loop. The algorithmic steps are as follows:
\begin{enumerate}
    \item Obtain a mini-batch sampled with the meta sampler from the training set and use that mini-batch to train a surrogate classification model. The surrogate model is used there to ensure the real model can be trained with a better batch from the meta sampler later. This step is done once per outer loop iteration.
    \item A loss for the surrogate network is computed on a class-balanced dataset with softmax and cross-entropy. The value of the loss was then used to update the meta sampler. This update is performed a few times per outer loop iteration.
    \item The improved sampler is then used to sample a new batch for the real network to train on. This step is done once per outer loop iteration.
\end{enumerate}
To enable end-to-end training, Ren et al.\cite{ren2020balsoftmax} used the reparameterization trick allowing the gradient to flow through the meta sampler.

Continuing with the logit adjustment framework, Menon et al. \cite{menon2021logitadjustment} propose a generalization of the logit adjustment methods. They observed that previous methods focused on scaling the logits and assumed the target distribution to be uniform. Instead, they propose a logit adjustment framework that can be implemented post-hoc to any existing model where an adjustment term based on the prior distribution of the target dataset is added to the logits. The adjustment can also be implemented as a loss term. The loss term would be adjusted based on added terms that would take into account the known prior of the target dataset. They showed that their method indeed was a generalization of previous methods and performed better on benchmark datasets.

Extending on the idea of logit adjustment, Zhang et al. \cite{zhang2021disalign} start by setting up and identifying an upper bound of the classification by including the test set in the dataset, making the test set known. They observed that the upper bound is not achieved by current methods when the test set is withheld. They propose a unified framework that learns a balanced network in 2 stages. First, they learn an adaptive calibration function that would linearly project frozen classifier logits into a balanced space. To determine how much calibration to use on the logit, they use a linear function with two learnable parameters that determine how much of the calibrated or the original logits to use based on the input data sample. This method effectively achieves both class-level and instance-level calibration of logits through a complete affine transformation of the logits. They then frame the imbalance classification problem as a distribution shift between an imbalanced training distribution and a balanced target distribution, with the objective of minimizing the KL divergence between the two distributions. The target distribution they conceived is a re-weighted version of the training distribution. The re-weighting is done by using the inverse of the class frequency so rare classes are weighted more than frequent classes.

Another interesting direction in this space was taken by Wang et al. \cite{wang2020ride}. They demonstrate that prior methods that tackle the long-tail problem by focusing on balancing the classifier head decrease tail bias at the expense of increased head bias, and increased variance for all classes. They propose an ensemble of expert models that are trained on balanced data. The experts are classifiers that share a backbone that learns the representations and are trained with a balanced loss based on the class frequency. The ensemble of experts benefits from the experts being sufficiently diverse from each other i.e. their posterior distributions over the classes are distant from one another. To make sure not too many experts are used, which would be demanding in resources, they use a router model that predicts whether to involve additional experts in the classification. The router model is trained to route to the next expert if the current expert is wrong but some $k$th expert correctly classifies the sample. They optionally apply self-distillation to distill the knowledge captured by many experts into fewer experts.

More than looking at the model performance on imbalanced classification, Zhong et al\cite{zhong2021mislas} also looked at the calibration error in the model's probability estimates over the classes. They observed that current methods have poor calibration of their output logits and are overconfident in their predictions. Usually by measuring the true accuracy against the predicted accuracy, the true accuracy is lower than the predicted accuracy. They also measured the Expected Calibration Error (ECE) and observed that the ECE is high for current methods (with lower being better), indicating that current methods have miscalibrated output logits. To address it, they first propose to use the Mixup\cite{zhang2018mixup} technique to generate synthetic samples for the rare classes. Then they also propose label smoothing of the classifier logits by artificially deducting some $\epsilon$ value on the correct class logit and redistributing that value to the other classes uniformly, effectively generating their own target posterior. $\epsilon$ is obtained as a function of the class sample count normalized by the difference between the maximum and minimum class sample counts. To fit the label-aware smoothing objective, they combine cRT\cite{kang2020decoupling} and LWS\cite{kang2020decoupling} into a single classifier update function that can either scale the classifier weights (LWS) or apply a learned update on the classifier weights (cRT) based on the hyperparameter retention factor $r$.

Continuing on the subject of miscalibrated output logits, Li et al. \cite{li2022ncl} tackle the issues by enlisting multiple experts. Each expert trains on a dataset with a balanced softmax. In addition to the normal view of the dataset, they select a set of hard classes where they define a hard class as a class that has the highest incorrect prediction score. The hard classes training, also with balanced softmax, is nested within the normal training. Those hard classes allow the model to focus on learning features that will help further discriminate them. In an attempt to exchange information between the experts, they train the experts to have similar posterior distributions (but not too similar as they found out experimentally). This is done by minimizing the KL divergence between the posterior distributions of the experts. However, if the experts are too similar, they are essential multiple instances of the same expert and the ensemble of experts will lose its diversity and benefits. To alleviate this, they keep the $\lambda$ parameter that controls how strongly those experts should be similar in the KL divergence loss term to not be too high (.6 was ideal in their experiments). By encouraging their experts to have the same posterior, they are, in an ad-hoc manner, encouraging the experts to distill their knowledge to one another. Finally, they add a supervised contrastive loss term to improve feature learning. The final prediction is obtained by the aggregation of the experts' output.

Looking at the weights of models trained on imbalanced datasets, Alshammari et al\cite{alshammari2022weightbal}
observed that imbalanced data caused the norm of the weights to be larger for the head classes than for the tail classes. This is because the head classes have far more samples than the tail classes. Ideally, the weights should be balanced so that the norms of the weights are the same for all classes. To tackle this issue, they propose a 2 stage training framework. In the first stage, they train the classifier with a weight decay term that penalizes the norm of the weights. In that first stage, they use cross entropy loss with a weight decay term. In the second stage, they train the classifier with weight decay and a max norm term that gives an upper bound to the weight norms. The max norm term is obtained by using the Lagrangian multiplier technique to convert the original constrained optimization problem into an unconstrained optimization problem. The new object is a min-max objective where the original loss is minimized by the classifier weights and the max norm term is maximized by the KKT multiplier\cite{wright2015coordinate} $\gamma$. Intuitively, as $\gamma$ grows larger, the second term dominates the objective, forcing the learned weights to be less than the upper bound $\delta$. To optimize both, the weights can be fixed to optimize $\gamma$, and then $\gamma$ can be fixed to optimize the weights. Combined with class-balanced sampling, the method achieves desirable uniform weight norms across all the classes.

\subsection{Representation Space}
The following works tackle imbalance in classification by exploiting the fact that representation learning is more robust to training on imbalanced distribution. The purple trapezoid displayed in Figure~\ref{fig:imbal-class} illustrates how this section fits in imbalance classification.

Kang et al. \cite{kang2021kcl} observed that the feature space learned by the model is imbalanced under long-tail distributions which, when paired with a classifier head, leads to poor decision boundaries for minority classes when using supervised cross-entropy loss. But, with contrastive loss, the feature space tends to be balanced. They then propose a supervised contrastive learning method that uses the labels of the data to learn a representation space where similar samples are close to one another and far from dissimilar samples. The supervised contrastive loss uses class labels to draw samples from the same class, also known as positive samples, close to one another while pushing samples from different classes, also known as negative samples, far from one another. They noticed however that, with a large number of samples, the majority classes would dominate the learning process. they then proposed their $K$-contrastive learning method (KCL) that limits the number of positive samples from each class to $K$ to prevent the frequent classes from dominating the learning process with more samples. They demonstrated their method also works on other tasks such as object detection and semantic segmentation.

However, a downside to contrastive methods is their need for a large number of negative samples. Wang et al.\cite{wang2021hybridsc} noticed the memory inefficiency in current supervised contrastive learning methods where they require large amounts of negative samples. To tackle this issue, they proposed a 2-branch network with a shared backbone. One branch is trained on imbalanced data samples with a Multi-Layer Perceptron to produce projections or embedding based on supervised contrastive loss. It leverages class prototypes (vectors representative of the class) for positive and negative pairs which are far fewer in numbers compared to data samples. Those class prototypes are learned as well. The other branch is trained on balanced data samples with a linear layer to produce good decision boundaries for the features based on supervised cross-entropy loss. The 2 branches are trained in a curriculum fashion with $\alpha$ controlling the balance between the 2 branches. $\alpha$ decays linearly with the epochs, indicating the training starts first with the feature learning branch and then gradually shifts to the classifier learning branch. More than one prototype per class can be used to improve the performance of the feature learning branch (still fewer in number compared to negative samples needed).

Zhu et al.\cite{zhu2022bcl} investigated the geometric structure formed by representation vectors of classes and the class prototypes. They observed that past methods fail to form regular simplex geometries in the feature space, which is crucial for the generalization of the learned representations. A regular simplex geometry has three crucial characteristics:
\begin{enumerate}
    \item The mean of all class prototypes should be the origin of the representation space.
    \item The class prototypes should be at a radius distance from the origin.
    \item The class prototypes should be vectors of which the dot product with one another can be calculated.
\end{enumerate}
Most methods fail to satisfy the first characteristic, which means that the mean of all class prototypes is not the origin and therefore the class prototypes are not at equal distance from each other. To address this they propose a 2 branch learning framework, where one branch is used to learn a feature extractor and a classifier called the classification branch. The other branch is used for contrastive learning called the contrastive learning branch. The classification branch employs cross-entropy loss and logit adjustment based on the class prior. The contrastive learning branch employs a novel loss called balanced contrastive loss (BCL) aiming to produce a regular simplex geometry in the feature space. The BCL is composed of two parts, a class averaging and a class component part. During experimentation, they observed that when trained on imbalanced data, regular Supervised Contrastive Learning (SCL), using the available examples for each class, grows the gradients for head classes far more than the tail classes. This is because the head classes have more examples than the tail classes. To alleviate that, class averaging works by averaging the instances of each class in a mini-batch so that each class has the same approximate contribution to the optimization. The class component part, which carries the bulk of the positive results obtained, consists of introducing a class prototype for each class in all the mini-batches. This allows all classes to be represented in all mini-batches and therefore all classes contribute equitably to the optimization. The class prototypes here are obtained from the MLP projection of the class-specific weights of the linear classifier in the classification branch and the system is trained end-to-end.

Li et al.\cite{li2022tsc} also noticed the importance of uniformity in the feature space and poor uniformity leads to poor separability in feature space. They proposed targeted supervised contrastive learning or TSC. They pre-process class targets to be uniform simplex geometry targets. They do this by generating uniform simplex geometry targets offline using stochastic gradient descent. Once the pre-computed targets are obtained, they assign each class to a target using the Hungarian algorithm to maintain the semantic meaning of the proximity between classes. In other words, if two classes are assigned to be close together, their features should be semantically similar. The targets are then used in the supervised contrastive learning loss to encourage the network to produce representations that match the target assignments and produce a uniform simplex geometry in the feature space. To understand learned representations, they propose 3 metrics: 
\begin{itemize}
    \item The intra-class alignment measuring how close the features of the same class are to each other (lower is better)
    \item The inter-class uniformity measuring the average distance between the centers of the classes (higher is better)
\end{itemize} 
However, the inter-class uniformity is not a good metric because it is invariant to the relative distance between neighboring classes and can therefore not measure if the class centers form a good simplex geometry. To alleviate this, they propose the 3rd metric, 
\begin{itemize}
    \item The neighborhood uniformity measuring the average distance between top $k$ neighboring classes (higher is better)
\end{itemize} 
They also introduce reasonability as a metric to measure the semantic meaning of the proximity between classes using WordNet hierarchy.

While some of the other methods address uncertainty in the predictions, Nam et al.\cite{nam2023srepr} turned their sight to uncertainty in the representations and observed that current methods have large uncertainty in their representations. They first observed that the Stochastic Weight Averaging (SWA) of the encoder weights at the end of the training (last 25\%) combined with the retraining of the classifier with the SWA weights resulted in improved performance compared to the absence of SWA. Furthermore, they can obtain the mean and covariance of the encoder weights to build a Gaussian distribution of the encoder weights estimating the posterior of the encoder weights, a technique called Stochastic Weight Averaging Gaussian (SWAG). Using SWAG, they can sample $M$ encoder weights from the Gaussian distribution over the weights and use them as a stochastic ensemble to produce $M$ representation of an input sample called stochastic representation. The stochastic representation captures the uncertainty in the representation space of the data. Using the classifier trained with the stochastic representation, they produce $M$ predictions for an input sample. The stochastic representation can be seen as the dispersion of the area of the representations in the feature space. A classifier trained on these stochastic representations will have better decision boundaries as it takes into account the uncertainty in the Representations. Once they complete the representation learning stage, they need to train the classifier. The naive option is to train the classifier that works on average with all the $M$ representations. However, they would like to capture the uncertainty in the predictions, not just the stochastic representations themselves. So, instead, they set up a distillation objective where the teachers are the $M$ classifiers with their encoders and the student, initialized as the SWA classifier, is further trained to maximally absorb the diversity in the teachers' predictions. Instead of directly distilling from the mean of the ensemble of prediction, they instead distill from the distribution of the predictions, so a distribution (Dirichlet) over the posterior distribution of the teachers' predictions over the classes. After training, they can take the mean of the student Dirichlet distribution as the final classification probability for an input sample.

\subsection{Non-uniform Test Distribution}
Almost all the works presented so far assumed that the test distribution for their proposed solution was balanced even though the training was imbalanced. However, that assumption is equally detached from reality where the test distribution in real-world applications has its own imbalance or is unknown altogether. The following works explore solutions to dealing with imbalance and unknown test distributions. This section is the far right side displayed in Figure~\ref{fig:imbal-class} where the Uniform Test Distribution in red is abandoned in favor of the Non-Uniform Test Distribution in green.

Hong et al.\cite{hong2021disentangling} do not assume that the target distribution is uniform but still assume it to be somehow known. Based on that, they re-frame the problem of imbalance classification as a distribution shift problem between the training distribution (imbalanced) and the test distribution. In their method, they disentangle the source distribution and logits of the model by having logits represent the ratio of the likelihood over evidence instead of directly representing the posterior. Given that the test distribution is known, they can obtain a balanced output posterior by using the test distribution prior and the network logits. They add a regularizer term to explicitly disentangle logits from source label distribution and use the known target label distribution to post-compensate for the imbalance during training.

Zhang et al. \cite{zhang2022sade} took a different approach where they don't assume the test distribution is known or uniform. They observed that current methods are not test distribution agnostic. Assuming the test distribution is unknown, they propose a self-supervised method to train a set of three experts on imbalanced, balanced, and reverse softmax loss respectively. Those experts are known as the forward expert, the uniform expert, and the reverse expert. The forward expert is trained on imbalanced (original) softmax loss with more focus on frequent classes, the uniform expert is trained on balanced softmax loss with more focus on middle classes, and the reverse expert is trained on reverse softmax loss with more focus on the rare classes. To intelligently aggregate the experts' logits, they train a linear aggregation model that combines the logits of each expert according to the learned weights associated with them. The aggregation function is learned by minimizing the prediction similarity between two augmented views of the same image passing through all the experts. This effectively leverages the stability of the experts where the more stable the expert is, the more it is weighted in the aggregation. Their method can handle not just uniform test distributions but also imbalanced and reverse test distributions.

\section{Imbalanced Regression}
\begin{figure*}[t!]  %
    \includegraphics[width=\textwidth]{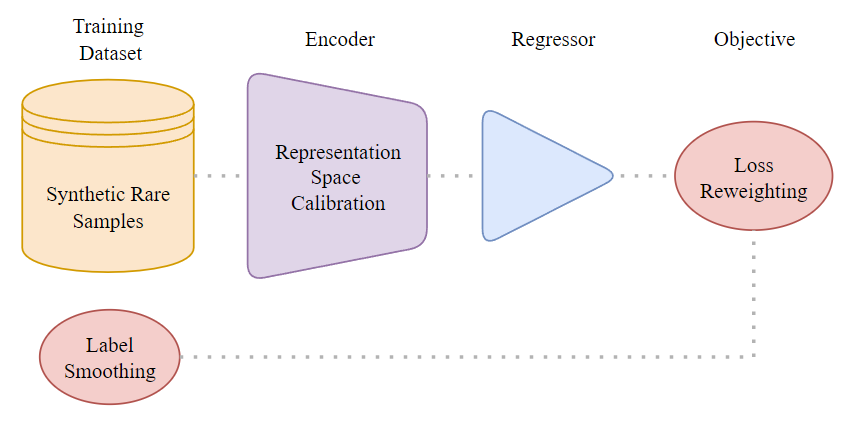} 
    \caption{Placement of the Imbalanced Regression groups in this section according to where they fit in the optimization process. The "Synthetic Rare Samples" works focus on adding to the training dataset. The "Label Smoothing" works focus on smoothing the training labels and is used in the objective. The "Representation Space Calibration" works focus on parameters of the encoder neural network. The "Loss Reweighting" works focus on the objective of the optimization process.}
    \label{fig:imbal-reg}
\end{figure*}

As one would naturally expect, imbalance also exists in regression tasks where, given a target continuous variable to predict, some ranges are far more frequent in the dataset compared to other ranges. Those ranges are called frequent or head cases, while the infrequent and non-existent ranges in the dataset are called rare or tail cases. The following works aim at tackling imbalance in regression from different angles with various methodologies. Figure~\ref{fig:imbal-reg} illustrates where those angles fit in the imbalanced regression.

\subsection{Synthetic Rare Samples}
To tackle the imbalance in regression, the following works rely on producing synthetic data samples for rare cases leveraging popular synthetic data generation frameworks. In Figure~\ref{fig:imbal-reg}, this section is illustrated by the far left cylinder in orange.

Bronco et al.\cite{branco2017smogn} proposes SMOGN which combines both under-sampling and over-sampling to tackle imbalance regression. First, the dataset is separated into two groups based on relevance, with high-relevance data being the target ranges where samples are rare and non-existent and low-relevance being the target ranges where samples are frequent. For irrelevant ranges, they perform under-sampling by randomly selecting a percentage of the frequent ranges' samples from the dataset. For relevant ranges, they combine two over-sampling techniques as follows: Given a rare case, they consider its $k$ nearest neighbors and calculate a safety distance as half of the median distance between the case and its neighbors. For all the $k$ neighbors that fall within the safety distance, they over-sample them by interpolating between them and the case. For neighbors falling into unsafe distances, they add a small Gaussian noise to the case to produce a synthetic sample. Both over-samplings are designed to produce very conservative synthetic data.

Moniz et al.\cite{moniz2018smoteboost}, in a similar spirit, propose SMOTEBoost which combines generating synthetic samples and boosting to tackle imbalance regression. First, they adapt and use the Synthetic Minority Oversampling Technique (SMOTE) for a prior work by Chawla et al.\cite{chawla2002smote} to generate new synthetic samples. With the augmented dataset, they use variants of AdaBoost to learn an ensemble of regressors. They use weak models at every iteration that learn from the mistakes of the prior iteration models as presented by Freund and Schapire in \cite{freund1995adaboost}.

\subsection{Label Space Smoothing}
Yang et al.\cite{yang2021ldsfds} noticed that, in imbalanced regression, the error distribution doesn't correlate well with the label distribution as it does in imbalanced classification. To tackle this issue, they apply a Gaussian kernel convolution operation to the label distribution to smooth it. The kernel size has to be not too small to avoid overfitting and not too large to avoid over-smoothing. With the right kernel size, each target range is smoothed with its neighboring ranges. They also explore the same process of smoothing for the feature space, which will be discussed in the next section 3.3. By itself, smoothing the labels is not sufficient to solve the imbalance. Their technique needs to be paired with some re-weighting based on, for example, the inverse of the smoothed label distribution. In Figure~\ref{fig:imbal-reg}, this section is illustrated by the far left red oval seating below the "Synthetic Rare Samples" cylinder because Label Smoothing is performed on top of the Training Dataset.

\subsection{Representation Space Calibration}
In this section, imbalanced in regression is tackled by first calibrating or smoothing the representations produced by the encoder neural network. The purple trapezoid in Figure~\ref{fig:imbal-reg} illustrates this section. 

As mentioned in the previous section, Yang et al.\cite{yang2021ldsfds} noticed that, in imbalanced regression, the error distribution doesn't correlate well with the label distribution as it does in imbalanced classification.
To tackle this issue, they apply a Gaussian kernel convolution operation to the label distribution to smooth it. However, they also observed that the feature or representation distribution is not well calibrated with the label distribution. To tackle this issue, they collect feature means and variances across bins. They then smooth those means and variances based on the neighboring bins' means and variances in the feature space. They calibrate the original features based on the smoothed means and variances obtained from the previous step through a distribution shift operation. To re-iterate the final point, their technique needs to be paired with some re-weighting based on, for example, the inverse of the smoothed label distribution.

Still focusing on Representation Space Calibration, Gong et al.\cite{gong2022ranksim} take a different approach by calibrating the feature space in such a way that, for any sample, the sorted list of its neighbors in the feature space is the same as the sorted list of neighbors in the target space. To achieve this, they compute the pairwise similarity matrices for both feature samples and label samples. They then compute the ranking similarity loss between the two matrices. Unfortunately for them, the rank function is not differentiable to allow end-to-end training. As a solution, they approximate the ranking function gradient by constructing a family of piece-wise affine continuous interpolations that trade off the informativeness of the gradient with fidelity to the original function. They then use the estimated gradient to update the model parameters. Their approach resulted in a mostly calibrated feature space.

\subsection{Loss Re-Weighting}
The following works directly tackle the imbalance in regression by re-weighting the loss function to nudge the model to focus more on the rare samples during training. In Figure~\ref{fig:imbal-reg}, it is shown as the red oval on the far right where it is related to the output of the regressor and the Label Smoothing applied on the Training Dataset. 

Steininger et al.\cite{steininger2021density} propose density-weighting to balance the loss function. First, they apply Gaussian kernel smoothing to the label distribution to estimate rare/missing ranges based on neighboring ranges similar to Yang et al.\cite{yang2021ldsfds}. They then obtain their density re-weighting function by taking the inverse of the smoothed label distribution and normalizing all data points' density values. Their density function has the following properties:
\begin{itemize}
    \item Samples with more common target values get smaller weights than rarer samples.
    \item Weighting function yields uniform weights for $\alpha = 0$, while larger $\alpha$ values further emphasize the re-weighting effect. 
    \item No data points are weighted negatively, as models would try to maximize the difference between estimate and true value for these data points during training.
    \item No weight should be 0 to avoid models ignoring parts of the dataset.
    \item The mean weight over all data points is 1. This eases applicability for model optimization with gradient descent as it avoids the influence of varying learning rates.
\end{itemize}
They apply the density weighting function as a coefficient to the regression loss function of choice to ensure the model focuses on rare or infrequent range samples.

Ren et al.\cite{ren2022bmse} also aim at re-weighting the loss function but solely focus on Mean Square Error (MSE). They obtained a balanced MSE by first assuming that the target distribution is uniform and doing a statistical transfer from a balanced uniform distribution to the original imbalanced distribution during training before testing the model on a balanced distribution. Unfortunately, the statistical transfer carries an integral term that is intractable to compute. To solve it, they explore 3 approaches to approximate the integral term: GAI, BMC, and BNI. GAI is GMM-based Analytical Integration that uses a Gaussian mixture of models to approximate the integral. BMC is Batch-based Monte Carlo that approximates the integral by an average over the batch. BNI is Bin-based Numerical Integration that approximates the integral by an average over the target bins. All three approaches yield acceptable approximations of the integral term, allowing them to train using their proposed Balanced MSE. 

\section{Representation Learning}
\begin{figure*}[t!]  %
    \includegraphics[width=\textwidth]{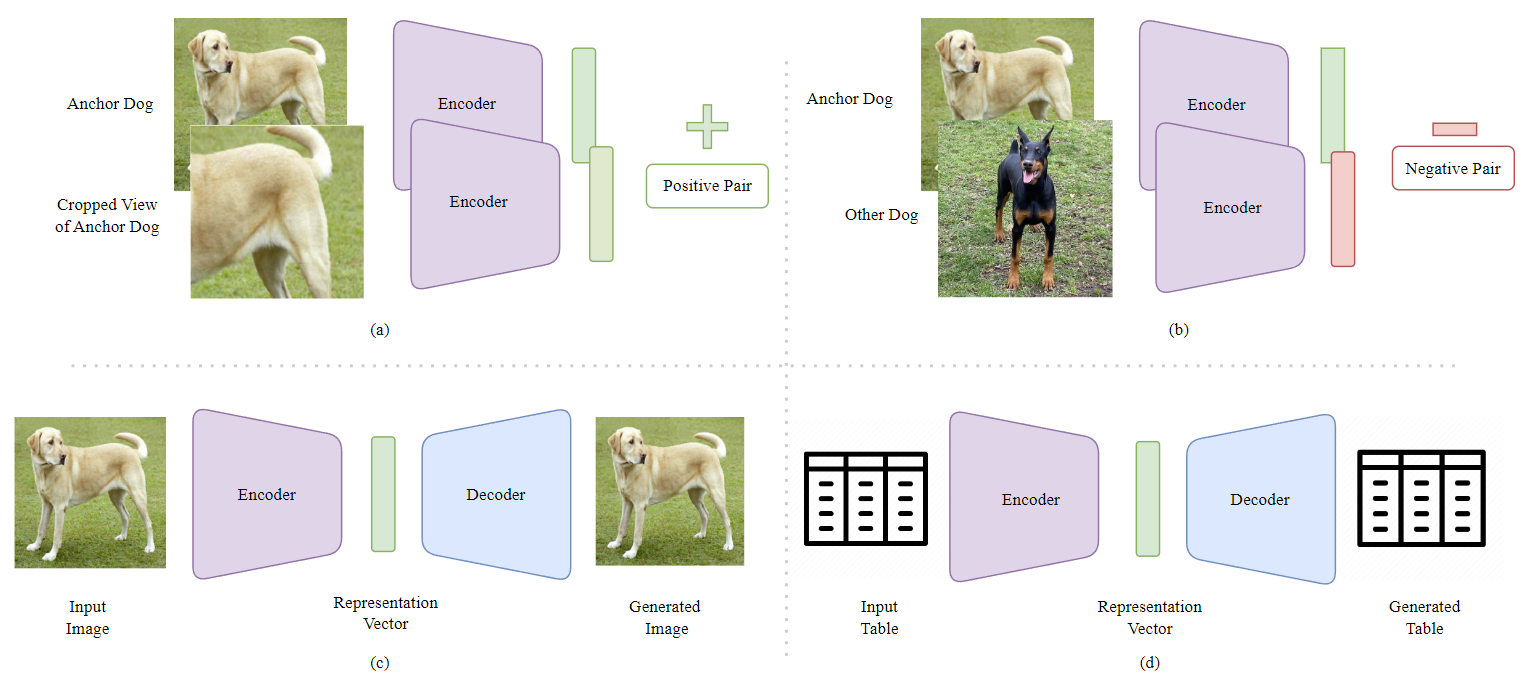} 
    \caption{(a) and (b) illustrate the contrast between Positive and Negative Pairs of Image samples used in Representation Learning. The anchor dog image forms a positive pair with the cropped view image and a negative pair with the image or view of another dog. Siamese networks are used to ingest the pair of views. On the other hand, (c) and (d) illustrate Representation Learning done with a decoder to generate or reconstruct the original input. In all cases, the input can be any modality, images in (a),(b), and (c), or tables in (d).}
    \label{fig:repr}
\end{figure*}

As we have seen in the previous sections, there was a wealth of research tackling imbalance distribution from many different angles for both classification and regression. Among the research works presented, a few of the more promising ones didn't just address the imbalance at the task level, whether be the classifier or the regressor model, but instead balanced the feature space which in turn provided representations and predictions that were robust to the imbalance. This set of works leveraged the properties of representation learning to handle imbalance. Below we will explore other works, not ones that directly tackle imbalance but ones that tackle representation learning, methods that have properties inherently robust to imbalance. Figure ~\ref{fig:repr} illustrates the different angles in Representation Learning.

\subsection{Positive and Negative Pairs}
Representation learning can be done in many ways. Here, we present works that tackle representation learning using both positive and negative pairs (illustrated in (a) and (b) Figure~\ref{fig:repr}). In representation learning, a positive pair is a pair of transformed (also known as augmented) views of the same input that should belong together and should be close to each other in feature space. A negative pair is a pair of transformed (augmented) views of different inputs that should not belong together and should be far from each other in feature space.

He et al.\cite{he2020moco} address representation learning with contrastive loss where an encoded query is matched to a dictionary of encoded keys. Contrastive methods (methods using contrastive loss) are usually sensitive to the number of negative examples, typically limited by the batch size. Compared to previous methods such as end-to-end training of two encoders or using a memory bank, this paper proposes a new method called Momentum Contrast (MoCo) that uses a queue to store multiple batches of encoded keys in FIFO style and a momentum update for the key encoder to ensure all encoded keys in the queue belong to the same representation (feature) space. The queue overcomes the limitations of end-to-end techniques which were limited by the overall hardware memory available for the batch and the momentum update ensures that the keys are in the same representation space, unlike the keys in the memory bank. From training, only the query encoder is updated by back-propagation. The key encoder is conservatively updated by momentum update where at most 10\% of the key encoder is updated by the query encoder. During experimentation, He et al.\cite{he2020moco} noticed Batch Normalization is not effective in this case because it leaks information via intra-batch communication across samples. To fix that, they introduce Shuffling Batch Normalization where they use multiple GPUs, performing batch normalization independently for each GPU and then shuffling the samples across GPUs.

Unlike He et al.\cite{he2020moco}, Chen et al.\cite{chen2020simclr} avoid the use of momentum update at the expense of larger batch size in their proposed technique called SimCLR. SimCLR uses a non-linear projection head on top of the encoder to get an embedding that is contrasted with other embeddings of the same sample. Given a sample, they apply a random composition of augmentations like crop, color, etc. to get a positive pair of augmented views from the same sample and contrast it with negative pairs of views from different samples in the batch. As such, to perform the best, the number of negative examples needs to be large, making the batch size large. They explored different combinations of augmentations and found that the best combination is a composition of random crop, color distortion, sobel filtering, and Gaussian blur. They introduced projection heads as simple Multi-Layer Perceptrons (MLP) before the contrastive loss to put distance between features and output so less information is lost in features. Overall, they found that introducing a SimCLR stage before fine-tuning the model on the desired downstream task improves the performance of the model.

Patacchiola and Storkey\cite{patacchiola2020relational} proposed a method that requires only one positive and one negative pair. They propose a relational reasoning framework with a relational reasoning module and a cross-entropy loss instead of using a contrastive loss. Their framework uses a relational network to ingest concatenated positive pairs from augmented views of the same sample and negative pairs from augmented views of different samples to output the relation probability of the pair. The relational network is trained with the binary cross-entropy loss. They explored different aggregation methods for the relational network and found that concatenation of the views is the best. Their method is more sample efficient compared to the other method needing greater numbers of negative pairs.

\subsection{Positive Pairs Only}
The works in this subsection, while still addressing representation learning, avoid the use of negative pairs altogether (illustrated in (a) Figure~\ref{fig:repr}). 

Grill et al.\cite{grill2020byol} proposes a framework called Bootstrap Your Own Latent (BYOL) that uses two branches, a momentum encoder, a projector head, a predictor head, and no negative pairs. BYOL works by producing two augmented views of the same sample and passing them to the two branches. The first branch has an online encoder updated by back-propagation producing a representation of the augmented view. That representation is passed to the projector head which is a MLP that outputs a projection of the representation. Finally, the projection is passed to the predictor head which is also an MLP that outputs a prediction of the projection of the other branch. The other branch has a target encoder updated by momentum update that produces a representation of the second augmented view. The representation is then passed to the projector head to get a projection of the representation. That second projection is the projection that the predictor head in the first branch is trained to output. The momentum update is at most 10\% update of the online encoder. BYOL uses symmetric loss, meaning that the augmented views are passed to both branches in the first order and then the reverse order. Grill et al. recognized that their approach was at risk of the trained model collapsing to a constant function where all inputs are mapped to the same representation. To avoid it, they use the momentum update on the target encoder branch.

Continuing with positive pairs only, Caron et al.\cite{caron2020swav} propose SwAV which doesn't match projections like the previous approaches but matches cluster assignments of augmented views. They start by learning cluster prototypes such that the cluster assignments of the positive pair of augmented views from the same sample are similar. Their framework consists of a two-branch network. From a single sample, they produce two augmented views and pass them to the two branches. First, the encoders of the 2 branches produce representations of the augmented views. The representations are passed into a projection head to produce projections of those representations. The projections are then passed into a cluster assignment head to produce cluster assignments of the projections to the cluster prototypes. The loss function compares the cluster assignment of one branch to the projection of the other and vice-versa symmetrically. To increase the performance of their approach, they also introduced multi-crop which consists of using two standard-resolution images and multiple low-resolution crops.

Observing the line of works in this domain, Chen and He\cite{chen2021simsiam} recognized that most proposed representation learning networks are Siamese Networks and that most tricks involved in training those networks are to avoid collapse to a constant representation. In an attempt to simplify Siamese Networks, they propose SimSiam which is a simple Siamese network that uses a predictor branch and a non-predictor branch with a stop gradient, with no need for negative pairs or momentum encoders. From a single sample, they generate two augmented views and pass them to the two branches in both this order and the flip order of the views (symmetric loss). Both branches have the same encoder, but only the predictive branch receives gradient updates. the encoders output representation and on the predictive branch, the representation is passed to a predictor head to predict the representation of the second branch. The loss is a symmetric cosine similarity loss between the prediction of the first branch and the projection of the second branch. They also show how their approach is similar to an expectation minimization algorithm and how important the stop-gradient is to prevent collapse.

Following in preventing collapse without negative pairs or momentum encoders, Zbontar et al.\cite{zbontar2021barlow} propose Barlow Twins leveraging the cross-correlation matrix between two augmented views of a samples batch. Given a batch of samples, they generate a random pair of batches of augmented views to be passed into a shared encoder and projection head to produce a pair of batches of projections of the batch of input samples. From the batches of projections, they calculate the cross-correlation matrix between the two batches of projections, crossing between a matrix (table) and another (table). Their objective for their cross-correlation matrix should be an identity matrix meaning that the same feature indices should be highly correlated (value of one) while different feature indices should be non-correlated (value of zero). To calculate the correlation, they assume the features have a mean at zero over the batch dimension and divide them by the largest cross-correlation value to avoid large feature values being interpreted as large correlations.

Bardes et al.\cite{bardes2021vicreg} propose a similar technique to Zbontar et al.\cite{zbontar2021barlow} called 
Variance, Invariance, and Covariance Regularizer (VICReg). The framework works by generating two augmented batches of views of the same input batch. The two batches are passed into an encoder and a projector to obtain two batches of projections. The first term in their loss is the invariance term which is the mean square distance between the two batches of projections. The invariance term serves to determine that the two batches of projections are similar since they are from the same batch of input samples. The second term is the variance term which is to prevent collapse to a constant function. The variance term ensures that the variance of the projections is above a threshold, meaning that there is enough variety between the projections and thus the representations are not collapsed to a constant vector. The third term is the covariance term which is to prevent information collapse. The covariance term ensures that the features of the projections are not correlated, meaning that the features are not functions of one or more features. In their ablative studies, they found that their methods perform the best when all 3 terms are used together with batch normalization.

Ermolov et al.\cite{ermolov2021whitening} took a slightly different approach where, instead of using siamese networks, they use a single branch network and the whitening process to prevent collapse. They first start by producing a lot more than one positive pair with no need for negative pairs by augmenting the input sample multiple times and passing the augmented views to the encoder and projector to produce multiple representations and then embeddings. They then whiten the embeddings by moving them to a zero mean one standard deviation Gaussian distribution. They normalize the embeddings so they rest in a unit circle. Then they calculate the MSE (similarity) between all positive pairs to attract them together. The whitening step requires calculating the $W_v$ matrix and $\mu$ mean of the embeddings. To better estimate the $W_v$ matrix, they partition the set of embeddings according to the augmentation applied. Use the same random permutation of elements across partitions to obtain sub-batches. They calculate $W_v$'s and $\mu$'s for the sub-batches repeating this process to obtain decent estimates of $W_v$'s and $\mu$'s.

\subsection{Generative Features}
All the approaches seen above rely on contrastive (discriminative) objectives to learn good features or representations. The following set of works instead relies on generative objectives such as reconstruction or masked prediction to learn generative features or representations (illustrated in (c) and (d) Figure~\ref{fig:repr}).

Wang et al.\cite{wang2022semantic} noticed that previous approaches to self-supervised learning or learning without labels for images relied on discriminative models to learn features. They proposed a framework called Semantic-aware Auto-encoders for Self-supervised Representation Learning where they use a generative model to help learn the features. The framework works by generating two augmented views of the same sample. One view is passed to the encoder to obtain a representation. The representation is passed to the decoder to attempt to reconstruct the other view. Unfortunately, the decoder cannot guess the other view from just the latent representation. So they add transformations on the encoder feature maps (representation with spatial information) to align with the transformations on the second view and pass the transformed feature maps to the decoder to produce the reconstructed image from which they obtain the final crop. They found that the transformations on the encoder feature maps are important for the decoder to learn. They also found That spatial information in feature maps for images is crucial and that global features that are not spatially aware are not good for reconstruction.

Instead of using augmentations, He et al.\cite{he2022mae} use masking in a framework they call Masked Auto-Encoders. They noticed that masking an image with random 16x16 patches covering more than 75\% of the image provides a difficult enough task for the encoder to learn a good semantic representation of the image without simply exploiting the spatial locality information in the image. In their Masked Auto-Encoders framework, they mask random 75\% patches of the image using 16x16 patches and pass only the visible patches to the encoder (Vision Transformer) to obtain a representation. They then add mask tokens to the representation, which are indicative of the absence of visible patches in those areas. The representation with the mask tokens is passed to the decoder to predict the missing patches. The authors proposed an asymmetric auto-encoder where the decoder is much smaller than the encoder to reduce computational cost. The decoder is then trained to predict only the missing patches. Through their experiments, they found that a high mask ratio offered a difficult enough task for the encoder to learn a good representation. They also found that passing only the visible patches reduced the computational cost of training a large encoder.

Following in the footsteps of He et al.\cite{he2022mae}, Xie et al.\cite{xie2022simmim} propose their SimMIM framework. Similarly to MAE, they Masked Auto-Encoders and masked a significant portion of the image with random 32x32 mask patches. However, unlike MAE, they passed both visible and mask patches to the encoder to obtain representations. Those representation vectors are passed to the decoder (without the need to pass mask tokens) to predict the missing patches. They went with a smaller decoder using a single linear layer decoder to predict the missing patches. Their training objective was to predict raw pixel values for the missing patches.

\subsection{Representation Learning on Tabular Data}
The modalities mostly addressed in representation learning are images and text. However, tabular data (illustrated in (d)  Figure~\ref{fig:repr}) are one of the most important modalities for real-world application of Deep Learning. The following works specifically address representation learning for tabular data.

Yoon et al.\cite{yoon2020vime} propose Value Imputation and Masked Estimation (VIME) with two stages, a self-supervised stage and a semi-supervised stage. In the self-supervised stage, they mask a random subset of features from a sample and then pass the obtained corrupted sample to the encoder which then outputs a representation. The representation is then passed to a decoder which outputs a reconstruction of the original sample and another decoder which outputs the mask applied to the samples. The second stage is a semi-supervised stage where they create several corrupted views of the same sample and pass it to the encoder along with the original sample. The encoder outputs a representation for each view. The representations are passed to the predictor's head to output predictions for the original samples and the corrupted views. The original sample prediction is compared to the original sample label and the corrupted views are compared with each other to make sure that, coming from the same original sample, they are consistent in predictions. In total, the network has three branches and four loss functions: the reconstruction and masked estimation loss for the first stage, and the Supervised and Consistency loss for the second stage. The masking process is done by randomly shuffling the values of the features (columns) of the samples in the batch.

Ucar et al.\cite{ucar2021subtab} propose Sub-setting features of tabular data (SubTab) for self-supervised representation learning. First, they divide an input instance into subsets. The subsets are then used to learn (subset) representations that, aggregated, form the representation of the input instance. During training, the subsets are corrupted with noise. First, the subsets are masked based on three masking schemes:
\begin{itemize}
    \item Random block of neighboring columns (NC)
    \item Random columns (RC)
    \item Random features per samples (RF)
\end{itemize}
Then, the masked features are going to be replaced by noise based on three noising schemes:
\begin{itemize}
    \item adding Gaussian noise
    \item overwriting a value of a selected entry with another one sampled from the same column
    \item zeroing-out selected entries
\end{itemize}
Once corrupted, the perturbed subsets were passed into a decoder to reconstruct the subsets or the original instance. Optionally, the paper proposed another branch with a projection head that would ingest the representation of the subsets and output a projection of the representation. Those representations would then be used to measure distance or similarity between subsets of the same instance and otherwise. For fine-tuning and inference, the subsets are not corrupted and the representations of the subsets are aggregated to form the representation of the instance.

Bahri et al.\cite{bahri2021scarf} propose a framework called Self-Supervised Contrastive Learning using Random Feature Corruption (SCARF) with two stages. The first stage is a self-supervised stage where they create a corrupted view of a data instance using a mask with some noise. They pass the instance and its perturbed view into an encoder and a projector to obtain the final projections for the instance and its corrupted view. Then they calculate the InfoNCE\cite{oord2018infoNCE} to find out the similarity between the instance and its corrupted view. The second stage is a supervised stage where they use the encoder from the first stage, remove the projector, and add a predictor head to predict the class of the instance. They compared many noising schemes for the perturbation and found that swap noise, or replacing the values of the features with values from other instances, is the best.

Somepalli et al.\cite{somepalli2022saint} noticed how deep learning methods are still behind traditional methods on tabular data. They proposed a framework called Self-Attention Inter-sample Attention Transformer (SAINT) for tabular data. First, they embed both categorical and numerical values in an embedding vector before passing it to their SAINT transformer. Their transformer uses two kinds of attention, self-attention between features and their proposed inter-sample attention between rows. The inter-sample attention is calculated by computing the attention score between the query row and all other rows. That attention score dictates how much of the features of other samples will be used to produce the representation of the query row. They also proposed a two-stage training where the first stage is pre-training in a self-supervised manner with a contrastive loss between the projection of the row and its noisy augmented view, and reconstructing the row from its noisy view's projection. The second stage is a supervised fine-tuning stage for the downstream task. They used CutMix\cite{yun2019cutmix} and MixUp\cite{zhang2018mixup} for data corruption.

Unlike the previous works that focus on learning good representations for tabular data, Gorishniy et al.\cite{gorishniy2022embeddings} aimed at addressing the embedding of numerical features in tabular data. 
Their framework is called Embeddings for Numerical Features in Tabular Deep Learning. In it, they propose two approaches for embedding numerical features in tabular data. The first approach is called Piece-wise Linear Encoding (PLE) where the numerical features are binned and each bin is encoded into a vector. Intuitively, the PLE represents how much the numerical value "fills" the embedding vector, where if the numerical value is greater than the bins' bounds, they are marked 1 in the target vector ("filled"), or the percentage of the bin filled if the value falls within the bin's bounds, or 0 otherwise ("not filled"). They discussed two approaches to finding the bins. The first is based on quantiles and the other is target-aware bins obtained by a decision tree. The second approach for embedding numerical features is called Periodic Position Encoding (PPE) where the numerical features are encoded into a vector of $sin$ and $cos$ curves of a source vector $v$ with $K$ learned coefficients of $2\pi x$ where $x$ is the numerical input. Intuitively, the PPE resembles a positional embedding for the value where $K$ represents the embedding dimension ($2K$ in total for $sin$ and $cos$), the range of the $C$ values represents the number of cycles of $2\pi$, the $C$s represents offsets from zero of the numerical value, and $x$ represents the multiplier of all $C$s to get the final value. Those $C$s were learned and saved as constants to be used in inference.

So far we have been exploring works that address imbalanced classification, imbalanced regression, and representation learning. Next, we are presenting a real-world application of deep learning that greatly suffers from imbalance and some of the prior works in this application.

\section{SEP Forecasting}
Solar Energetic Particles (SEPs) are high-energy, charged particles emitted from the Sun, often accelerated by solar flares or coronal mass ejections (CMEs). These events are frequently accompanied by X-ray and radio flux bursts, which can serve as early indicators of imminent SEP activity. Such accompanying phenomena are often monitored through space weather observatories to provide a more comprehensive understanding of the solar events triggering SEPs. These particles pose a significant risk to both unmanned and manned space missions; they can damage electronic circuits in spacecraft and induce acute radiation sickness in astronauts. Accurate forecasting of SEPs is therefore critical, allowing for timely preventive measures like system shutdowns, spacecraft reorientation, or even the delay of launches and spacewalks. Understanding and mitigating the impact of SEPs is not just a matter of operational efficiency, but a vital safety requirement for the advancement of space exploration. However, when applying current deep learning methods (usually assuming balance in the data) on SEP  forecasting, they do much worse. The reality is that SEP events are much rarer than the regular background noise of space. The datasets collected by NASA on the Sun show extreme imbalance where often only 1\% of the dataset are SEP events. Whether taking into account the imbalance or only considering a balance sandbox of the dataset, below are works that 
apply machine learning to SEP forecasting.

\begin{figure*}[t!]  %
    \includegraphics[width=\textwidth]{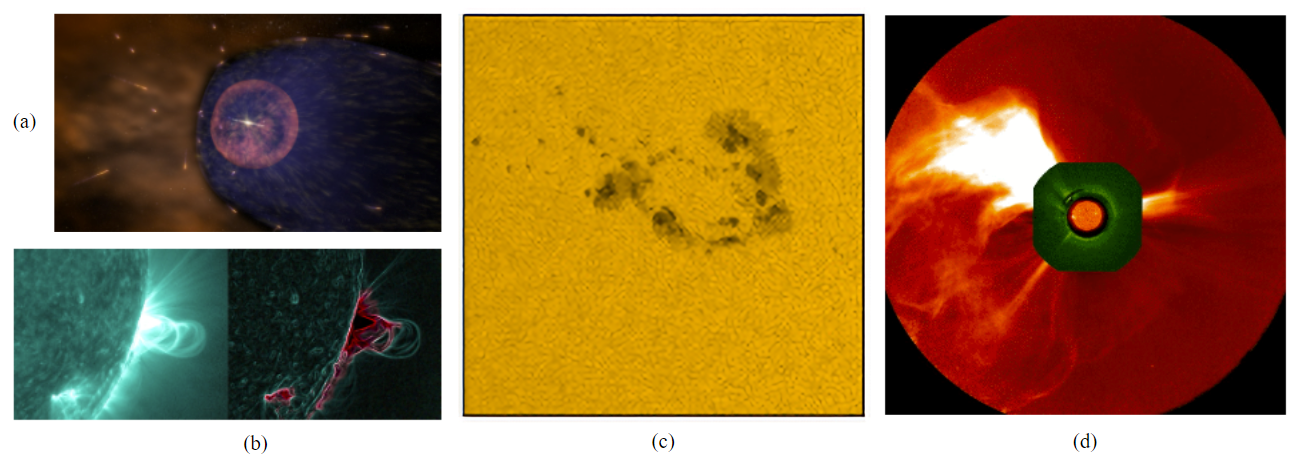} 
    \caption{Illustrations of Solar Events used in works within this section for SEP forecasting. Front the left to the right, we have (a) Solar X-rays, (b) Solar Fluxes, (c) Active Regions, and (d) CMEs.}
    \label{fig:solar-events}
\end{figure*}

\subsection{Primarily CME Features}
The following works all share some use of Coronal Mass Emissions (CMEs) data to forecast SEPs. Figure~\ref{fig:solar-events}(d) offers an illustrative view of a CME.

Richardson et al.\cite{richardson2018prediction} analyzed the effectiveness of the Richardson Equation (Richardson et al. 2014) for predicting the intensity of 14-24MeV protons in SEPs using CME speed and connection angle as features. The Richardson equation which is $$ I(\phi)(MeV\cdot s \cdot cm^2 \cdot sr)^{-1} \approx 0.013 \cdot exp(0.0036V - \frac{\phi^2}{2\sigma^2})) $$, $\sigma$ = 43° where the 43° Gaussian width is the average obtained by Richardson et al. (2014). They found that the Richardson equation leads to a lot of false alarms compared to true predictions of SEPs. By increasing the threshold for the intensity considered SEP, they found that the false alarm rate reduces but so does the number of SEPs they can actually catch. They then explore using the CME speed for filtering out false alarm rates which yields the same reduction in false alarm rate at the cost of missing SEPs as well. They explore using the CME width as well resulting in a variant of the same trade-off. They explore using interplanetary and non-interplanetary type 2 Radio emissions, weak, moderate, and bright Type 3 Radio emissions power and duration as additional filters. They found these filters reduce the false alarm rate but miss weaker SEPs. They conclude that the Richardson equation is a decent basis for the empirical prediction of SEPs.

Torres \cite{Torres2020ml} tackled SEP forecasting using Proton, Electrons, and X-ray time-series data. Their approach takes as input electron and proton intensities 2 hours in the past to output proton intensity in 30 minutes and an hour in the future. They also add phase information to every 5-minute timestep as additional input features. The phases are the onset, threshold, peak, and end periods. They applied a Multi-Layer Perceptron (MLP) and Recurrent Neural Network (RNN) to the forecasting task. They divide their forecasting into 3 levels where a model (MLP or RNN) is dedicated to training and testing on that level only. To decide which of the levels is appropriate for each model, they employ either a routing algorithm based on predefined thresholds with the models only focusing on the intensity level associated with them or use a phase selection model to route the time series intensities to different phase models only focusing on the phase level associated with them. They add X-rays to the electron data as additional features. To measure the performance of the approaches, they measure the Mean Absolute Error between the actual and predicted intensities, the lag between the prediction and the actual intensity forecast for the on-set to peak, and for the onset to the threshold, to measure the detection of the rising edge and of the precursory intensity signal respectively. Finally, they measure the difference between when the target and when the prediction reaches reach ln(10) each. Their experiments showed that the RNN did slightly better than the MLP at forecasting from their time series data.

Lavasa et al.\cite{lavasa2021assessing} addressed the prediction of SEPs using CME and Solar Flare features. They start with an initial sample analysis where they find that for about 33,221 Solar Flare events, 6218 are CMEs and only 257 are SEPs, illustrating the extreme imbalance in the data. Before applying any model to the data, they preprocess it and extract 8 features, 6 Solar Flare features, and 2 CME features. They then normalize the features or apply the logarithm to make sure that the values and ranges of those features are manageable. They apply a Nested Cross Validation scheme where there are inner folds are used for finding hyperparameters for the models and outer folds for testing the found hyperparameters. It works by first splitting the dataset into $k$ folds. They hold on out for testing and use the $k-1$ folds for training. Of the $k-1$ folds, they perform another $n$-folding where they divide the combined $k-1$ folds into $n$ folds. They hold one out for validation and train on the remaining $n-1$ folds. They train 8 machine learning including Linear models, SVMs, Neural Nets, and Decision Tree algorithms compare their relative performance in balanced and imbalanced settings. No clear winner can be declared as the algorithms trade wins over different metrics. They analyze the importance of the given features in the classification, finding out that Solar Flare fluence, CME speed, and width are the most useful to predict SEPs.

Griessler \cite{griessler2023rtae} addressed the forecasting of SEPs from non-SEPs and the intensity of the SEPs. For classifying SEPs using CME features, they explore 3 methods, oversampling the minority classes to obtain a more balanced dataset, applying classifier retraining (cRT\cite{kang2020decoupling}), and an auto-encoder head to learn better features during the first stage (cRT+AE). For cRT+AE, they estimate the coefficient contribution of the auto-encoder loss to ensure both branches are equally valued in the training. They found cRT+AE to perform the best in their experiments. For predicting the SEP natural logarithm intensity, they explore 6 methods, oversampling the minority ranges, retraining the regressor (rRT), adding an auto-encoder head to learn better features during the first stage (rRT+AE), integrating the Richardson equation \cite{richardson201425} prediction by linear combination or by using the neural network to predict an adjustment error term to the Richardson prediction and applying dense loss \cite{steininger2021density} reweighting to the loss function so the training model pays more attention to rare samples. They evaluate their performance using Mean Absolute Error and Pearson Correlation Coefficient. They found that rRT+AE with either dense loss \cite{steininger2021density} or Richardson \cite{richardson201425} adjustment error yielded the best results.

\subsection{Primarily X-Ray Features}
The following works all share some use of X-ray data to forecast SEPs. Figure~\ref{fig:solar-events}(a) offers an illustrative view of Solar X-rays.

Boubrahimi et al.\cite{boubrahimi2017satellite} tackle the problem of predicting $>$100 MeV SEP events from a time series of proton and x-ray channels with a decision tree. They first start by analyzing X-ray and proton channels as multivariate time series that entail some correlation which may be precursors to the occurrence of SEP. They study the correlation across all channels. In total, they have 8 channels: xs for short wavelength x-ray (.5 -.3 nm), xl for long wavelength x-ray (.1 - .8 nm), and 6 proton channels $p6_flux$ to $p11_flux$ with different energetic intervals from 80MeV to  $>$700MeV. To analyze the correlations in channels, they consider the span and the lag in their observation. The span is the number of hours that constitute the observation period prior to an X-ray event. The lag is the factor by which a value of a time series is multiplied to produce its previous value in time. To express the X-ray and proton cross-channel correlations, they used a vector auto-regression model (VAR) where the dependent variables are the proton channels at $t$ and the independent variables are the proton and x-ray channels at time previous times determined by the lag. After fitting the VAR to the data, they use the coefficient as features fed to a Decision Tree to predict SEP events based on the assumption that the coefficient values would increase as precursors to the event.

Kahler and Ling\cite{kahler2018forecasting} address the SEP forecasting with flare X-ray peak ratios. They found that flares associated with >10MeV SEPs of >10 proton flux units have lower peak temperatures than those without SEPs. They first analyze how the ratios of the short (0.05 - 0.4nm) and long (0.1 - 0.8nm) wavelengths x-rays peak fluxes and peak flare fluxes of long wavelengths are useful features for distinguishing SEP events of greater intensity against lower energy and background events. They further analyzed the Eastern and Western hemisphere events and found that the Western hemisphere events are better classified by the peak-flux ratios and peak fluxes. Using the peak flux ratio between the short and long wavelength and the peak fluxes of the long wavelength x-ray as features, they train an MLP and a K-Nearest Neighbor algorithm to classify high-intensity SEPs from low-intensity and background events.

Aminalragia-Giamini et al.\cite{aminalragia2021solar} address the prediction of SEP events using short and long wavelength X-rays and 24 Solar Flare features such as cosine and sine of the heliolongitude, peak flux, and fluence, etc. For their classifier, the authors used an ensemble of 3 MLPs whose outputs are averaged. They analyze the relationship between X-ray peak fluxes in the long band and the number of SEPs in the dataset to find that More SEPs are related to more intense X-rays. They artificially balanced the dataset and found that their method achieved good performance on the balanced dataset. To further analyze the performance of their model, they focused on type M2 solar flares and found that the uncertainty associated with Solar Flares of M2 and greater classes is larger in their method. They analyzed the decision threshold of their method and found it to be best at around .79 for Solar Flares of class C1 and above, and .5 for solar Flares of class M2 and above.

\subsection{Primarily Active Region Features}
The following works all share some use of Active Regions, which are high-magnetism regions, to forecast SEPs.  Figure~\ref{fig:solar-events}(c) offers an illustrative view of an Active Region.

Inceoglu et al.\cite{inceoglu2018using} address predicting whether Solar Flares will be associated with CME and SEPs. First, they start by comparing the number of events associated with Solar Flares alone compared to solar flares with CME and SEPs. They found that Solar Flares associated with CMEs and SEPs tend to be of higher classes. They also compared the number of events associated with the speed of CMEs when they are alone and when they are with Solar Flares and SEPs. They also found that CMEs associated with Solar Flares and SEPs tended to have higher speeds. They also consider Active Regions which are regions of high magnetism. Active regions have been known to generate SEPs. The authors considered features from active regions at some time in the past to predict current SEP events. They employed a Support Vector Machine (SVM) and an MLP to ingest those features to output the right class of events. They found that the SVM tended to be more accurate than the MLP.

Kasapis et al.\cite{kasapis2022interpretable} address SEPs forecasting using SMARP Features and Solar Flare features. They first analyze Active Regions (AR) which are regions of high magnetic fields, and how these regions are related to Solar Flares and SEPs. They track 5 AR properties and 2 Solar Flare properties which are used as inputs to a Linear Model and an SVM. They train and test their models using balanced as well as imbalanced data, where the balancing is done by oversampling the rare samples. They found that they obtained the best performance by leveraging a subset of features from SMARP and Solar Flares that were the most discerning of SEP events. They found that the flares features performed better than the SMARP Active Region features. They also observed that the performance of the algorithms would drop significantly as the level of imbalance in training and testing increased.

\subsection{Primarily Solar Radio Flux Features}
The following work mainly relied on Solar Radio Flux (SRF) to forecast SEPs.  Figure~\ref{fig:solar-events}(b) offers an illustrative view of Solar Radio Fluxes.

Kim et al.\cite{kim2018technique} address the prediction of SEP events using Solar Radio Flux. By looking at Solar Radio Flux (SRF) at 2800MHz, 1415MHz, and 610MHz, they analyze the solar activity related to those fluxes and the SEPs that might be related. First, they analyzed the probability distribution on the relative overall rate of increase of SRF related to SEPs and found that a significant increase in the rate of SRF was correlated to an increase in the number of SEPs. They also looked at the probability distribution of daily total SRF related to SEPs and found that an increase in the SRF level was correlated to an increase in the number of SEPs when only looking at SEPs. Moreover, the 610MHz band seems to be the earliest and highest correlated precursor to SEP events. Using these findings, they pass all 3 bands' SRF rate of increase and the daily total SRF level as features to a Neural Network to predict the number of SEPs. They propose to find the neural network's optimal weights using gradient descent or a Genetic Algorithm. Finally, they propose a second neural network that is tasked to predict SRF to generate more recent data (within 3 days) if it's not available so it can be used by their primary model for SEP forecasting.

\section{Conclusion}
Deep Learning has catalyzed advancements across numerous domains, from image recognition to weather forecasting. Nevertheless, a pressing issue remains the incongruity between lab-based training assumptions and real-world data distributions. This is particularly evident in applications like weather prediction, where low-incidence but high-impact events like storms are not adequately captured by current models.

The scientific community has responded by innovating training approaches and designing architectures that are more attuned to imbalanced datasets. One promising avenue is representation learning, which enhances model generalizability by facilitating a more sophisticated understanding of the feature space. This is especially pertinent for mitigating the limitations imposed by data imbalance.

SEP forecasting presents a challenging test case for these advancements. The high stakes associated with this application, coupled with the rarity of specific SEP events, make it an ideal benchmark for evaluating the real-world efficacy of these methods. 

In summary, the landscape of Deep Learning presents a multitude of open challenges that span imbalanced classification, imbalanced regression, and representation learning. These challenges range from addressing data intrinsic characteristics and developing robust sampling techniques in imbalanced scenarios to designing disentangled and interpretable representations. Furthermore, these issues become even more complex and urgent in specialized applications like Solar Energetic Particle (SEP) forecasting, where data quality and model reliability are of paramount importance. Most of the methods applied in that space do not address the imbalance in a significant way yet. Addressing these challenges will be crucial for bridging the gap between theoretical advancements and real-world applicability, opening new avenues for groundbreaking research and practical innovations.

\bibliographystyle{plain}
\bibliography{references}

\end{document}